\documentclass[journal]{IEEEtran}
\usepackage{graphicx}
\usepackage{amsmath,amssymb} 
\usepackage{multirow}

\ifCLASSINFOpdf
\else
\fi
\newcommand{\tabincell}[2]{\begin{tabular}{@{}#1@{}}#2\end{tabular}}

\begin{document}
%
\title{Robust LSTM-Autoencoders for Face De-Occlusion in the Wild}
%
%
%

\author{\IEEEauthorblockN{Fang Zhao, Jiashi Feng, Jian Zhao, Wenhan Yang, Shuicheng Yan
		\thanks{Fang Zhao, Jiashi Feng, Jian Zhao and Shuicheng Yan are with Department of Electrical and Computer Engineering, National University of Singapore, Singapore, e-mail: \{elezhf, elefjia\}@nus.edu.sg, zhaojian90@u.nus.edu, eleyans@nus.edu.sg.}%
		\thanks{Wenhan Yang is with Institute of Computer Science and Technology, Peking University, Beijing, 100080, P.R. China, e-mail: yangwenhan@pku.edu.cn.}
	}
	
}

%
%

\markboth{IEEE Transactions on Image Processing, Draft}%
{Shell \MakeLowercase{\textit{et al.}}: Bare Demo of IEEEtran.cls for IEEE Journals}
%



\maketitle

\begin{abstract}
Face recognition techniques have been developed significantly in recent years. However, recognizing faces with partial occlusion is still challenging for existing face recognizers which is heavily desired in real-world applications concerning surveillance and security. Although much research effort has been devoted to developing face de-occlusion methods, most of them can only work well under constrained conditions, such as all the faces are from a pre-defined closed set. In this paper, we propose a robust LSTM-Autoencoders (RLA) model to effectively restore partially occluded faces even in the wild. The RLA model consists of two LSTM components, which aims at occlusion-robust face encoding and recurrent occlusion removal respectively. The first one, named multi-scale spatial LSTM encoder, reads facial patches of various scales sequentially to output a latent representation, and occlusion-robustness is achieved owing to the fact that the influence of occlusion is only upon some of the patches. Receiving the representation learned by the encoder, the LSTM decoder with a dual channel architecture reconstructs the overall face and detects occlusion simultaneously, and by feat of LSTM, the decoder breaks down the task of face de-occlusion into restoring the occluded part step by step. Moreover, to minimize identify information loss and guarantee face recognition accuracy over recovered faces, we introduce an identity-preserving adversarial training scheme to further improve RLA. Extensive experiments on both synthetic and real datasets of faces with occlusion clearly demonstrate the effectiveness of our proposed RLA in removing different types of facial occlusion at various locations. The proposed method also provides significantly larger performance gain than other de-occlusion methods in promoting recognition performance over partially-occluded faces.
\end{abstract}


%
\IEEEpeerreviewmaketitle

\section{Introduction}

In recent years, human face recognition techniques have demonstrated promising performance in many large-scale practical applications. However, in real-life images or videos, various occlusion can often be observed on human faces, such as sunglasses, mask and hands. The occlusion, as a type of spatially contiguous and additive gross noise, would severely contaminate discriminative features of human faces and harm the performance of traditional face recognition approaches that are not robust to such noise. To address this issue, a promising solution is to automatically remove facial occlusion before recognizing the faces \cite{R1, R2, R3, R5, R6}. However, most of existing methods can only remove facial occlusions well under rather constrained environments, e.g., faces are from a pre-defined closed set or there is only a single type of occlusion. Thus those methods are not applicable for the complex real scenarios like surveillance.

\begin{figure}[t]
	\setlength{\abovecaptionskip}{0pt}
	\centering
	\includegraphics[height=6cm]{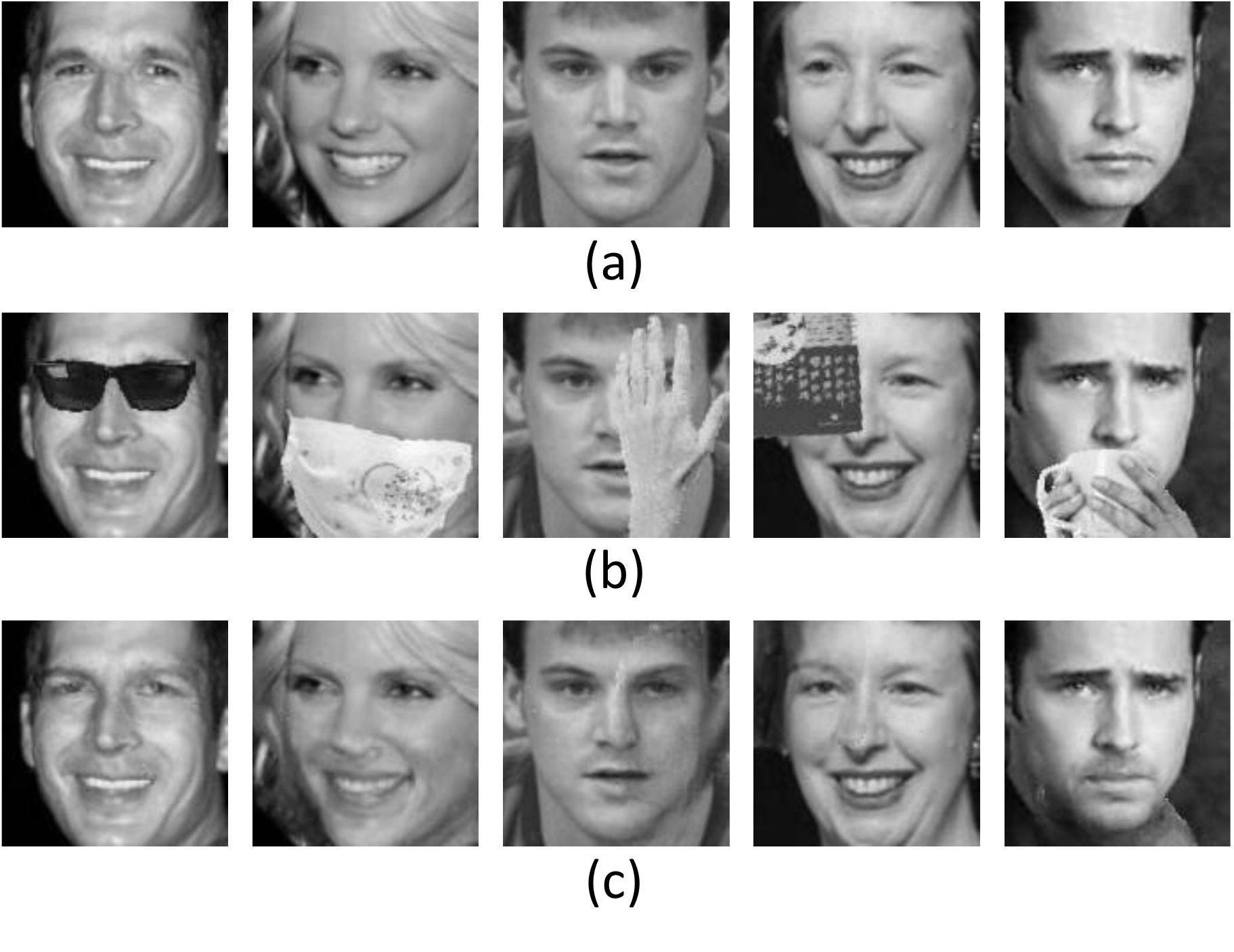}
	\caption{We address the task of face de-occlusion with \textbf{various types of occlusions} under the condition of \textbf{open test sets} (i.e., test samples have no identical subject with training samples). (a) Original occlusion-free faces; (b) Occluded faces; (c) Recovered faces by our proposed method.}
	\label{fig:show_example}
\end{figure}

In this work, we aim to address this challenging problem -- face de-occlusion \emph{in the wild} where the faces can be from an open test set and the occlusions can be of various types (see Fig.~\ref{fig:show_example}). To solve this problem, we propose a novel face de-occlusion framework built upon our developed robust LSTM-Autoencoders (RLA). In real scenarios, facial occlusion often presents rather complex patterns and it is difficult to recover clean faces from the occluded one in a single step. Different from existing methods pursuing one-stop solution to de-occlusion, the proposed RLA model removes occlusion in several successive processes to restore occluded face parts progressively. Each step can benefit from recovered results provided by the previous step. More concretely, the RLA model works as follows.

Given a new face image with occlusion, RLA first employs a multi-scale spatial LSTM encoder to read patches of the image sequentially to alleviate the contamination from occlusion in the encoding process. RLA produces a occlusion-robustness latent representation of the face because the influence of occlusion is only upon some of the patches. Then, a dual-channel LSTM decoder takes this representation as input and jointly reconstructs the occlusion-free face and detects the occluded regions from coarse to fine. The dual-channel LSTM decoder contains two complementary sub-networks, i.e. a face reconstruction network and an occlusion detection network. These two networks collaborate with each other to localize and remove the facial occlusion. In particular, hidden units of the reconstruction network feeds forward the decoding information of face reconstruction at each step to the detection network to help the occlusion localization, and the detection network back propagates the occlusion detection information into the reconstruction network to make it focus on reconstructing occluded parts. Finally, the reconstructed face is integrated with the occluded face in an occlusion-aware manner to produce the recovered occlusion-free face. We train the overall RLA in an end-to-end way, through minimizing the mean square error (MSE) between paired recovered face and ground truth face. We observe that purely minimizing MSE usually over-smoothes the restored facial parts and leads to loss of the discriminative features for recognizing person identity. This would hurt the performance of face recognition. Therefore, in order to preserve the identity information of recovered faces, we introduce an identity based supervised CNN to encourage RLA to preserve the discriminative details during face recovery. However, this kind of supervised CNN results in severe artifacts in the recovered faces. We thus further introduce an adversarial discriminator \cite{R24}, which learns to distinguish recovered and original occlusion-free faces, to remove the artifacts and enhance visual quality of recovered faces. As can be seen in the experiments, introducing such discriminative regularization indeed effectively preserves the identity information of recovered faces and facilitates the following face recognition.

Our main contributions include the following three aspects. 1) We propose a novel LSTM autoencoders to remove facial occlusion step by step. To the best of our knowledge, this is the first research attempt to exploit the potential of the LSTM autoencoders for face de-occlusion in the wild. 2) We introduce a dual-channel decoding process for jointly reconstructing faces and detecting occlusion. 3) We further develop a person identity diagnostic de-occlusion model, which is able to preserve more facial details and identify information in the recovered faces through employing a supervised and adversarial learning method.

\section{Related Work}

\subsection{Face De-Occlusion}

There are some existing methods based on analytic--synthetic techniques for face de-occlusion. Wright et al. \cite{R1} proposed to apply sparse representation to encoding faces and demonstrated certain robustness of the extracted features to occlusion. Park et al. \cite{R2} showed that eye areas occluded by glasses can be recovered using PCA reconstruction and recursive error compensation. Li et al. \cite{R3} proposed a local non-negative matrix factorization (LNMF) method to learn spatially localized and part-based subspace representation to recover and recognize occluded faces. Tang et al. \cite{R5} presented a robust Boltzmann machine based model to deal with occlusion and noise. This unsupervised model uses a multiplicative gating to induce a scale mixture of two Gaussians over pixels. Cheng et al. \cite{R6} introduced a stacked sparse denoising autoencoder with two channels to detect noise through exploiting the difference between activations of the two SSDAs, which requires faces from training and test sets have the same occluded location. All of those methods do not consider open test sets. Test samples in their experiments have the identical subjects with training samples, which is too limited for practical applications.

\subsection{Image Inpainting}

Our work is also related to image inpainting which mainly aims to fill in small image gaps or restore large background regions with similar structures. Classical image inpainting methods usually is based on local non-semantic algorithms. Bertalmio et al. \cite{R25} proposed to smoothly propagate information from the surrounding areas in the isophotes direction for digital inpainting of still images. Criminisi et al. \cite{R26} introduced a best-first algorithm to propagate the confidence in the synthesized pixel values in a manner similar to the propagation of information in inpainting and compute the actual colour values using exemplar-based synthesis. Osher et al. \cite{R27} proposed an iterative regularization procedure for restoring noisy and blurry images through using total variation regularization. It is difficult for those methods to remove gross spatially contiguous noise like facial occlusion because too much structural information is lost in that case, e.g., the entire eye or mouth is occluded.

Recently, some methods based on global context features have been developed. Xie et al. \cite{R15} proposed the stacked sparse denoising autoencoders (SSAD) for image denoising and inpainting through combining sparse coding and pre-trained deep networks. Pathak et al. \cite{R29} trained the context encoders to generate images for inpainting or hole-filling and simultaneously learned feature representations which captures appearances and semantics of visual structures. However, the locations of image regions which require to be filled in are provided beforehand. By contrast, our method dose not need to know locations of the corrupted regions and automatically identify those regions.

\section{Robust LSTM-Autoencoders for Face De-Occlusion}

In this section we first briefly review the Long Short-Term Memory (LSTM). Then we elaborate the proposed robust LSTM-Autoencoders in details, including the multi-scale spatial LSTM encoder, the dual-channel LSTM decoder and the identity preserving component.

\begin{figure*}[t]
	\centering
	\includegraphics[height=7.5cm]{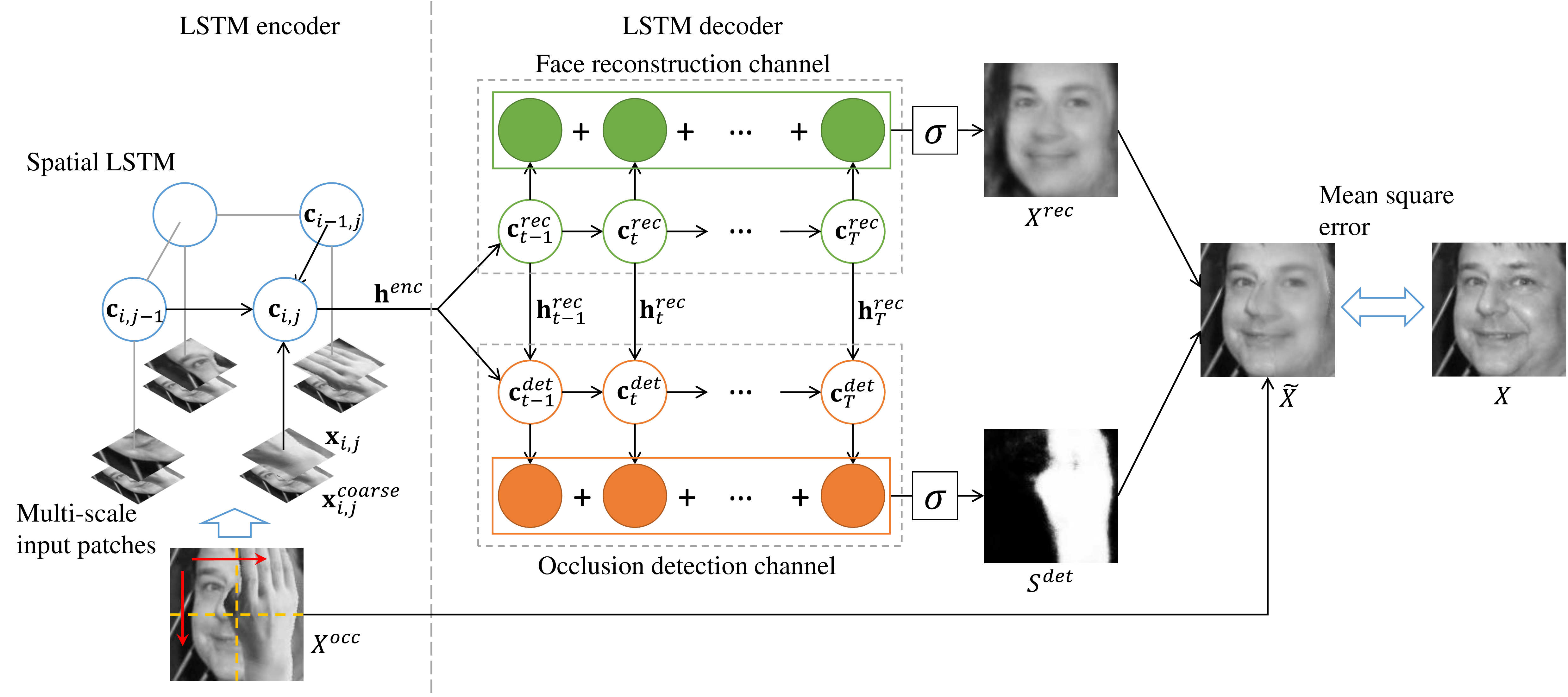}
	\caption{Illustration on the framework of our proposed robust LSTM-Autoencoders and its training process. It consists of a multi-scale spatial LSTM encoder and a dual-channel LSTM decoder for concurrent face reconstruction and occlusion detection.}
	\label{fig:framework}
\end{figure*}

\subsection{Long Short-Term Memory}

Long Short-Term Memory (LSTM) \cite{R7} is a popular architecture of recurrent neural networks. It consists of a memory unit $\bf{c}$, a hidden state $\bf{h}$ and three types of gates --- the input gate $\bf{i}$, the forget gate $\bf{f}$ and the output gate $\bf{o}$. These gates are used to regulate reading and writing to the memory unit. More concretely, for each time step $t$, LSTM first receives an input ${\bf{x}}_t$ and the previous hidden state ${\bf{h}}_{t - 1}$, then computes activations of the gates and finally updates the memory unit to ${\bf{c}}_{t}$ and the hidden state to ${\bf{h}}_{t}$. The involved computation is given as follows,
\begin{align}\label{1}
	& {{\bf{i}}_t} = \sigma ({{\bf{W}}_{xi}}{{\bf{x}}_t} + {{\bf{W}}_{hi}}{{\bf{h}}_{t - 1}} + {{\bf{b}}_i}), \nonumber \\
	& {{\bf{f}}_t} = \sigma ({{\bf{W}}_{xf}}{{\bf{x}}_t} + {{\bf{W}}_{hf}}{{\bf{h}}_{t - 1}} + {{\bf{b}}_f}), \nonumber \\
	& {{\bf{c}}_t} = {{\bf{f}}_t} \odot {{\bf{c}}_{t - 1}} + {{\bf{i}}_t} \odot \tanh ({{\bf{W}}_{xc}}{{\bf{x}}_t} + {{\bf{W}}_{hc}}{{\bf{h}}_{t - 1}} + {{\bf{b}}_c}),  \\
	& {{\bf{o}}_t} = \sigma ({{\bf{W}}_{xo}}{{\bf{x}}_t} + {{\bf{W}}_{ho}}{{\bf{h}}_{t - 1}} + {{\bf{b}}_o}), \nonumber \\
	& {{\bf{h}}_t} = {{\bf{o}}_t} \odot \tanh ({{\bf{c}}_t}) \nonumber
\end{align}
where $\sigma(x) = 1 / (1 + \exp (-x))$ is a logistic sigmoid function, $\odot$ denotes the point-wise product, and $\mathbf{W}$ and $\mathbf{b}$ are weights and biases for the three gates and the memory unit.

A major obstacle in using gradient descent to optimize standard RNN models is that the gradient might vanish quickly during back propagation along the sequence. LSTM alleviates this issue effectively. Its memory unit sums up activities over all time steps, which guarantees that the gradients are distributed over the factors of summation. Thus, back propagation would not suffer from the vanishing issue anymore when applying LSTM to long sequence data. This makes LSTM memorize better long-range context information.

Due to such an excellent property, LSTM has been extensively exploited to address a variety of problems concerning sequential data analysis, e.g., speech recognition \cite{R8}, image captioning \cite{R9}, action recognition \cite{R10} and video representation learning \cite{R12}, as well as some problems that can be casted to sequence analysis, e.g., scene labeling \cite{R13} and image generation \cite{R14}. Here we utilize LSTM netwroks to build our face de-occlusion model where facial occlusion is removed by a sequential processing to eliminate the effect of occlusion step by step.

\subsection{Robust LSTM-Autoencoders}

In this work, we are going to solve the problem of recovering a occlusion-free face from its noisy observation with occlusion. Let $X^{occ}$ denote an occluded face and let $X$ denote its corresponding occlusion-free face. Face de-occlusion then aims to find a function $f$ that removes the occlusion on $X^{occ}$ by minimizing the difference between the recovered face $f(X^{occ})$ and the occlusion-free face $X$:
\begin{equation}\label{2}
	\min_f \left\| {f({X^{occ}}) - X} \right\|_F^2.
\end{equation}
We propose to parameterize the recovering function $f$ using an autoencoder, which has been exploited for image denoising and inpainting \cite{R15}. The recovering function then can be expressed as
\begin{equation}\label{3}
	f({X^{occ}}) = {f_{dec}}({f_{enc}}({X^{occ}};{\bf{W}},{\bf{b}});{\bf{W'}},{\bf{b'}}),
\end{equation}
where $\{\bf{W},\bf{b}\}$ and $\{\bf{W'},\bf{b'}\}$ encapsulate weights and biases of the encoder function and decoder function respectively. In the image denoising and inpainting, the goal is to remove distributed noise, e.g., Gaussian noise, and contiguous noise with low magnitude, e.g., text. Unlike them, one cannot apply the autoencoder directly to remove facial occlusion. It is difficult to remove such a large area of spatially contiguous noise like occlusion in one step, especially in unconstrained environments where face images probably have various resolutions, illuminations, poses and expressions, or even never appear in training data. Inspired by divide-and-conquer algorithms \cite{R23} in computer science, here we propose an LSTM based autoencoder to divide the problem of de-occlusion into a series of sub-problems of occlusion detection and removal.

Fig.~\ref{fig:framework} illustrates the framework of our proposed robust LSTM-Autoencoders (RLA) model. We now proceed to explain each of its components and how they work jointly to remove facial occlusion one by one.

\subsubsection{Multi-scale Spatial LSTM Encoder}

Given the architecture shown in Fig.~\ref{fig:framework}, we first explain the built-in LSTM encoder. The LSTM encoder learns representations from the input occluded face $X^{occ}$. Here it is worth noting that if the LSTM encoder takes the whole face as a single input, the occlusion will be involved in the overall encoding process and eventually contaminate the generated representation. In order to alleviate the negative effect of occlusion, as shown in the left panel of Fig.~\ref{fig:framework}, we first divide the face image into $M \times N$ patches, denoted as $\{ {{\bf{x}}_{i,j}}\} _{i,j = 1}^{M,N}$ (here $M=N=2$), and feed them to a spatial LSTM network sequentially. Spatial LSTM is an extension of LSTM for analyzing two-dimensional signals \cite{R16}. It sequentializes the input image in a pre-defined order (here, from left to right and top to bottom). By doing so, some of encoding steps will see occlusion-free patches and thus not be affected by noise. Besides, the noisy information from occluded patches is not directly encoded into feature representations, but controlled by the gates of spatial LSTM for the sake of the subsequent occlusion detection. At each step, the LSTM also encodes a larger region ${\bf{x}}_{i,j}^{coarse}$ around the current patch but with a lower resolution to learn more contextual information. Here the whole image is used as ${\bf{x}}_{i,j}^{coarse}$ and concatenated with ${\bf{x}}_{i,j}$ as a joint input of the encoder.

For each location $(i,j)$ in the $M\times N$ grid dividing the image, the multi-scale spatial LSTM encoder learns representations from the patch centered at $(i,j)$ as follows,
\begin{align}\label{4}
	& \left( {\begin{array}{*{20}{c}}
			{{{\bf{i}}_{i,j}}}\\
			{{{\bf{f}}_{i - 1,j}}}\\
			{{{\bf{f}}_{i,j - 1}}}\\
			{{{{\bf{\tilde c}}}_{i,j}}}\\
			{{{\bf{o}}_{i,j}}}
		\end{array}} \right) = \left( {\begin{array}{*{20}{c}}
		\sigma \\
		\sigma \\
		\sigma \\
		{\tanh }\\
		\sigma
	\end{array}} \right){F_{{\bf{W}},{\bf{b}}}}\left( {\begin{array}{*{20}{c}}
	{{{\bf{x}}_{i,j}}}\\
	{{\bf{x}}_{i,j}^{coarse}}\\
	{{{\bf{h}}_{i - 1,j}}}\\
	{{{\bf{h}}_{i,j - 1}}}
\end{array}} \right), \\
& \ \ {{\bf{c}}_{i,j}} = {{\bf{f}}_{i - 1,j}} \odot {{\bf{c}}_{i - 1,j}} + {{\bf{f}}_{i,j - 1}} \odot {{\bf{c}}_{i - 1,j}} + {{\bf{i}}_{i,j}} \odot {{{\bf{\tilde c}}}_{i,j}}, \nonumber \\
& \ \ {{\bf{h}}_{i,j}} = {{\bf{o}}_{i,j}} \odot \tanh ({{\bf{c}}_{i,j}}), \nonumber
\end{align}
where $F_{{\bf{W}},{\bf{b}}}$ is an affine transformation w.r.t. parameters $\{\bf{W},\bf{b}\}$ of the memory unit and gates respectively (ref. Eqn.~\eqref{1}). The memory unit ${\bf{c}}_{i,j}$ is connected with two previous memory units ${\bf{c}}_{i-1,j}$ and ${\bf{c}}_{i,j-1}$ in the 2-D space. It takes the information of neighboring patches into consideration when learning the representation for the current patch.

After reading in all patches sequentially, the spatial LSTM encoder outputs its last hidden state ${\bf{h}}_{M,N}$ in the sequence as a feature representation ${\bf{h}}^{enc}$ of the occluded face. The representation is then recurrently decoded to extract face and occlusion information for face recovery.

\subsubsection{Dual-Channel LSTM Decoder}

Given the representation ${\bf{h}}^{enc}$ of an occluded face produced by the encoder, an LSTM decoder follows to map the learned representation back into an occlusion-free face. Traditional autoencoders, which have been used in image denoising, usually perform the decoding for once only. However, as we explain above, faces may contain a variety of occlusion in the real world. This kind of spatially contiguous noise corrupts images in a more malicious way than general stochastic noise such as Gaussian one, because it incurs loss of important structural information of faces. As a result, the face cannot be recovered very well by only one-step decoding. Therefore, we propose to use an LSTM decoder to progressively restore the occluded part.

As shown in the top right panel of Fig.~\ref{fig:framework}, the LSTM decoder takes over ${\bf{h}}^{enc}$ as its input ${\bf{h}}_{0}^{rec}$ at the first step and initializes its memory unit with the last memory state of the encoder ${\bf{c}}^{enc}$, and then keeps revising the output ${X_t}$ at each step $t$ based on the previous output ${X_{t - 1}}$. The operations of the LSTM decoder for face reconstruction can be summarized as
\begin{align}\label{5}
	& \left( {\begin{array}{*{20}{c}}
			{{\bf{i}}_t^{rec}}\\
			{{\bf{f}}_t^{rec}}\\
			{{{\bf{\tilde c}}}_t^{rec}}\\
			{{\bf{o}}_t^{rec}}
		\end{array}} \right) = \left( {\begin{array}{*{20}{c}}
		\sigma \\
		\sigma \\
		{\tanh }\\
		\sigma
	\end{array}} \right){F_{{\bf{W}},{\bf{b}}}^{rec}}({{\bf{h}}_{t - 1}^{rec}}), \\
& \ \ {{\bf{c}}_t^{rec} = {\bf{f}}_t^{rec} \odot {\bf{c}}_{t - 1}^{rec} + {\bf{i}}_t^{rec} \odot {\bf{\tilde c}}_t^{rec},}\\
& \ \ {{\bf{h}}_t^{rec} = {\bf{o}}_t^{rec} \odot \tanh ({\bf{c}}_t^{rec}),}\\
& \ \ {X_t = X_{t - 1} + {\bf{W}}^{rec}{\bf{h}}_t^{rec} + {\bf{b}}^{rec},}
\end{align}
where $rec$ indicates that the parameters are used for the reconstruction network. The final reconstructed face ${X^{rec}} = \sigma ({X_T})$  is obtained by passing the output at the last step $T$ through a sigmoid function, which can be seen as a result refined by decoding for multiple times.

In the above decoding and reconstruction process, we apply the decoder on both non-occluded and occluded parts. Thus, pixels of non-occluded parts may suffer from the risk of being corrupted in the decoding process. To address this issue, we introduce another LSTM decoder which aims to detect the occlusion. Being aware of the location of occlusion, one can simply compensate values of the non-occluded pixels using original pixel values in the inputs. In particular, for each pixel, the occlusion detector estimates the probability of its being occluded. As illustrated in Fig.~\ref{fig:framework} (bottom right), for each step $t$, the LSTM detection network receives the hidden state ${\bf{h}}_t^{rec}$ of the reconstruction network, and updates its current occlusion scores ${S_t}$ based on the previous detection result. Here the cross-network connection provides the decoding information of face reconstruction at each step for the detection network to better localize the occlusion. More formally, the LSTM decoder detects occlusion as follows,
\begin{align}\label{6}
	& \left( {\begin{array}{*{20}{c}}
			{{\bf{i}}_t^{det}}\\
			{{\bf{f}}_t^{det}}\\
			{{\bf{\tilde c}}_t^{det}}\\
			{{\bf{o}}_t^{det}}
		\end{array}} \right) = \left( {\begin{array}{*{20}{c}}
		\sigma \\
		\sigma \\
		{\tanh }\\
		\sigma
	\end{array}} \right){F_{{\bf{W}},{\bf{b}}}^{det}}\left( {\begin{array}{*{20}{c}}
	{{\bf{h}}_t^{rec}}\\
	{{\bf{h}}_{t - 1}^{det}}
\end{array}} \right), \\
& \ \ {{\bf{c}}_t^{det} = {\bf{f}}_t^{det} \odot {\bf{c}}_{t - 1}^{det} + {\bf{i}}_t^{det} \odot {\bf{\tilde c}}_t^{det},}\\
& \ \ {{\bf{h}}_t^{det} = {\bf{o}}_t^{det} \odot \tanh ({\bf{c}}_t^{det}),}\\
& \ \ {{S_t} = S_{t - 1} + {\bf{W}}^{det}{\bf{h}}_t^{det} + {\bf{b}}^{det},}
\end{align}
where $det$ indicates that the parameters are used for the detection network. Similar to the reconstruction network, the final occlusion scores are given by ${S^{det}}= \sigma ({S_T})$. Then combining the reconstructed face $X^{rec}$ and the occluded face $X^{occ}$ according to the occlusion scores $S^{det}$ gives the recovered face ${\tilde X}$ with compensated pixels:
\begin{equation}\label{7}
	\tilde X  = {X^{rec}} \odot {S^{det}} + {X^{occ}} \odot (1 - {S^{det}}).
\end{equation}
Note that ${\tilde X}$ is actually a weighted sum of $X^{rec}$ and $X^{occ}$ using $S^{det}$. The pixel value in the reconstructed face $X^{rec}$ is fully preserved if the score is one, and the pixel value is equal to the one from the occluded face $X^{occ}$ if its occlusion score is zero.

\begin{figure*}[t]
	\centering
	\includegraphics[height=5.5cm]{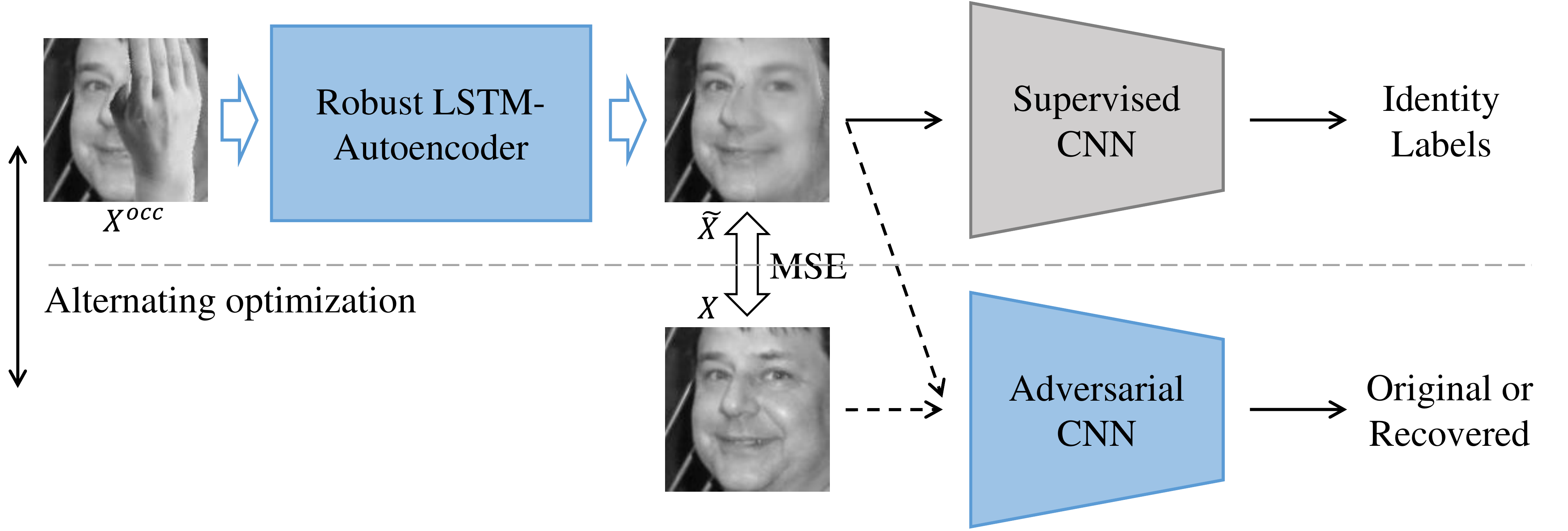}
	\caption{Pipeline of identity-preserving RLA with the supervised and adversarial CNNs. The gray box indicate that parameters of the network are fixed during the fine-tuning stage. MSE: mean square error.}
	\label{fig:supervised_learning}
\end{figure*}

\subsubsection{Optimization}

Given a training dataset $\{ X_i^{occ},{X_i}\} _{i = 1}^K$, substituting Eqn.~(\ref{7}) into Eqn.~(\ref{2}), we have the following mean square error function that RLA is going to optimize
\begin{multline}\label{8}
	{L_{mse}}({\bf{W}},{\bf{b}}) = \\
	\frac{1}{{2K}}\sum\limits_{i = 1}^K {\left\| {X_i^{rec} \odot S_i^{det} + X_i^{occ} \odot (1 - S_i^{det}) - {X_i}} \right\|_F^2},
\end{multline}
which can be minimized by standard stochastic gradient descent. Taking its derivatives w.r.t. $X_i^{rec}$ and $S_i^{det}$ gives the gradients:
\begin{align}\label{9}
	& \frac{{\partial L}}{{\partial {X_i^{rec}}}} = \frac{1}{{K}}(\tilde X_i - X_i) \odot {S_i^{det}}, \\
	& \frac{{\partial L}}{{\partial {S_i^{det}}}} = \frac{1}{{K}}(\tilde X_i - X_i) \odot ({X_i^{rec}} - {X_i^{occ}}).
\end{align}
Then they are used in error back propagation to update the parameters of each LSTM network. Note that in Eqn.~(\ref{9}), the gradients according to the non-occluded part are set to zeros by the occlusion sores $S^{det}$, and thus the reconstruction network will prefer to reconstruct the occluded part with the help of the occlusion detection network.

Since the model contains three networks, i.e., the encoder network, the face reconstruction network and the occlusion detection network, directly training the three networks simultaneously hardly gives a good local optimum and may converge slowly. To ease the optimization, we adopt a multi-stage optimization strategy. We first ignore parameters of the occlusion detection network, and pre-train the encoder and decoder to minimize the reconstruction error $\sum\nolimits_{i = 1}^K {\left\| {X_i^{rec} - {X_i}} \right\|_F^2}$. Then we fix their parameters and pre-train the decoder for occlusion detection to minimize the joint loss in Eqn.~\eqref{8}. These two rounds of separate pre-training provides us with sufficiently good initial parameters and we proceed to retrain all the three networks jointly. We observe that this strategy usually gives better results and faster convergence rate in the experiments.

\begin{figure}[t]
	\centering
	\includegraphics[height=3.2cm]{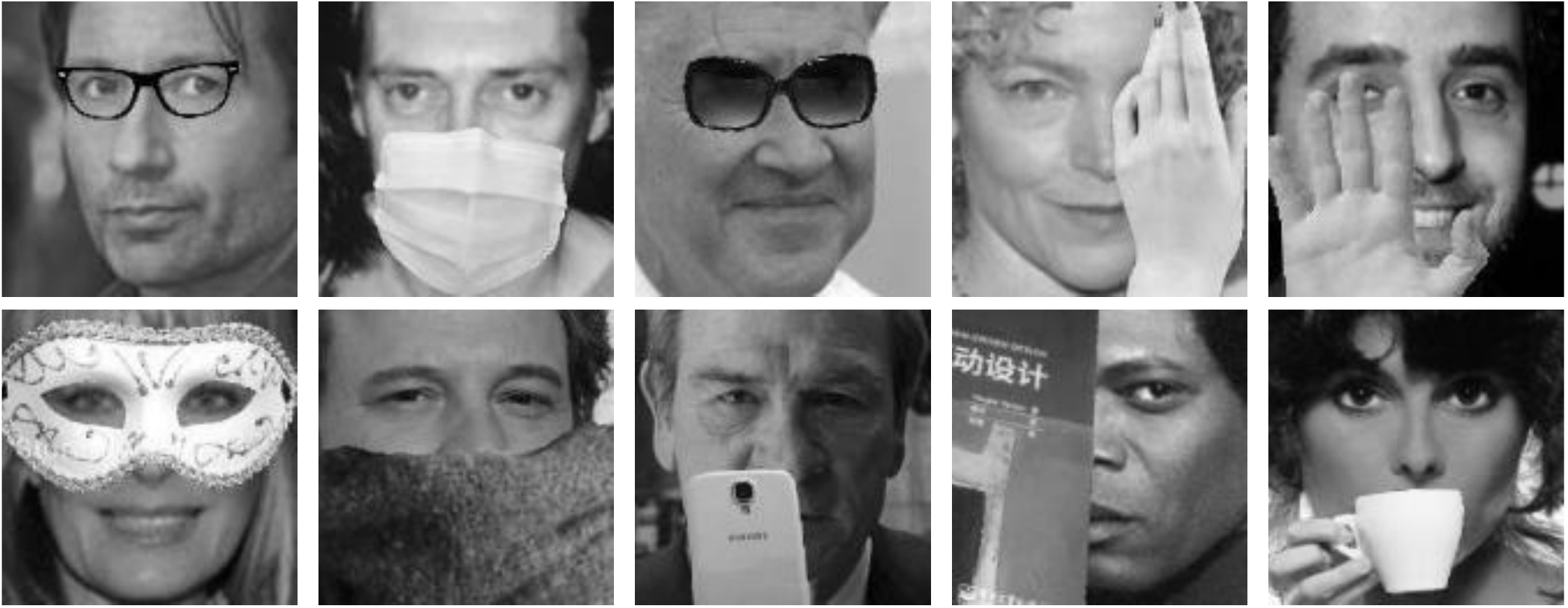}
	\caption{Samples from the occluded CASIA-WebFace dataset which is used for training. In total 9 types of occlusion (50 templates for each type) are synthesized, including sunglasses, masks, hands, glasses, eye masks, scarfs, phones, books and cups.}
	\label{fig:occulded_examples}
\end{figure}

\subsection{Identity-Preserving Face De-Occlusion}

Although it is able to restore facial structural information (e.g., eyes, mouth and their spatial configuration) from occluded faces well, the RLA model introduced above only considers minimizing the mean squared error between occlusion-free and recovered faces. Generally, there are multiple plausible appearances for an occluded facial region. For example, when the lower face is occluded, only according to the upper face, it is hard to determine what the lower face is actually like. Thus if we force the model to exactly fit the value of each pixel, it would tend to generate mean values of all probable appearances for the recovered part. This probably leads to loss of discriminative details of faces and harms the performance of face recognition. Recently, deep convolutional neural networks (CNNs) are widely applied to face recognition and provide state-of-the-art performance \cite{R17,R18,R19}. Inspired by their success, we propose to leverage an identity based supervised CNN and an adversarial CNN to provide extra guidances for the RLA model on face recovery, in order to preserve the person identity information and enhance visual quality of recovered faces.

\begin{figure*}[t]
	\centering
	\includegraphics[height=10cm]{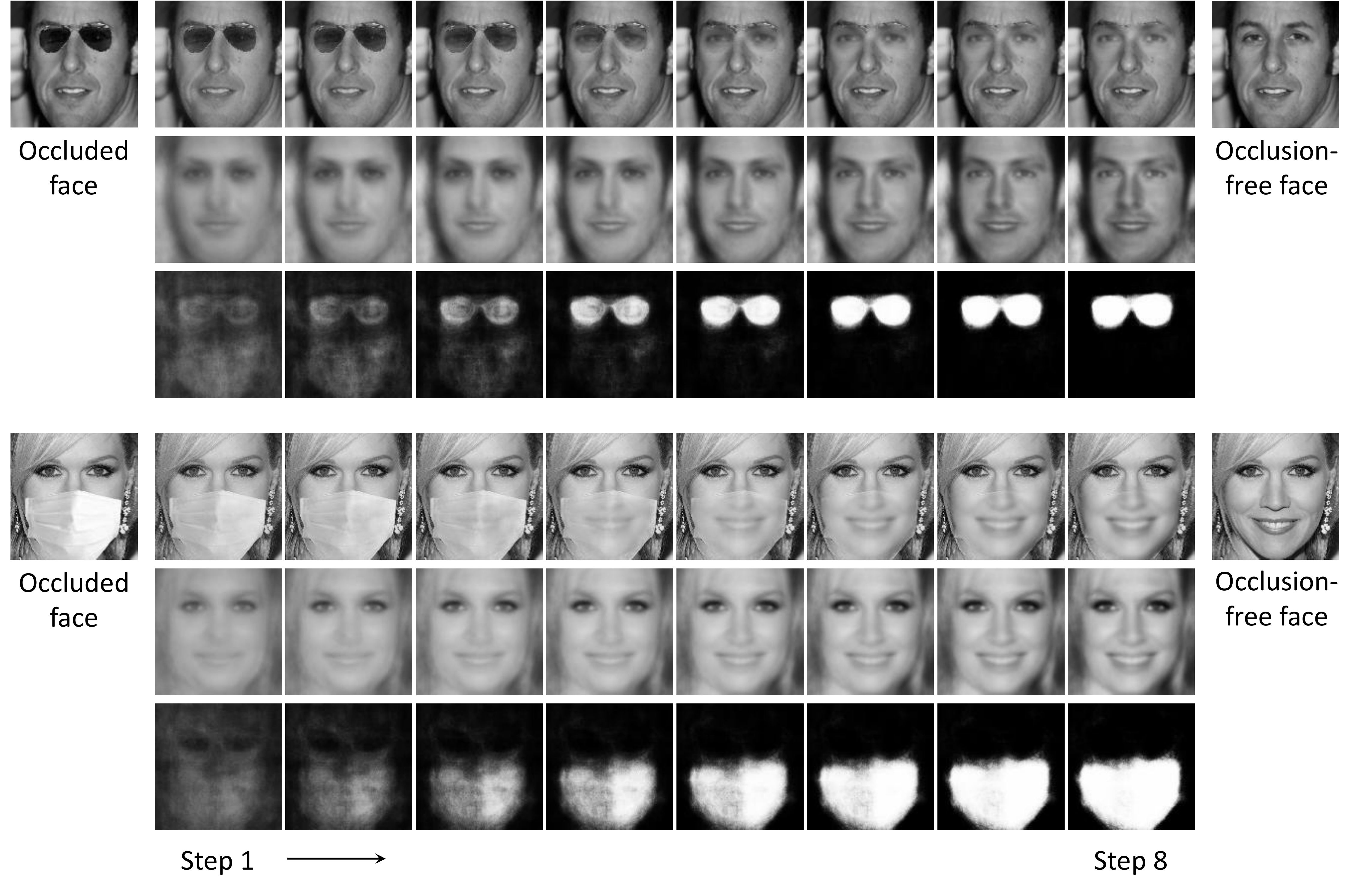}
	\caption{Occlusion removal process of RLA. The middle columns display outputs of RLA from step 1 to 8. The first row shows outputs of the decoder with both face reconstruction and occlusion detection components. The second row shows outputs from only face reconstruction (without occlusion detection). The third row displays the detection results of occlusion.}
	\label{fig:results_step}
\end{figure*}

Fig.~\ref{fig:supervised_learning} illustrates our proposed pipeline for identity-preserving RLA (IP-RLA). A pre-trained CNN is concatenated to the decoder for classifying recovered faces with identity labels $\{y_i\}_{i = 1}^K$, and helps tune RLA to simultaneously minimize the mean squared error between pixels ${L_{mse}}$ in Eqn.~(\ref{8}) and the classification loss
\begin{equation}
{L_{sup}} = \frac{1}{K}\sum\limits_{i = 1}^K { - \log (P({y_i}|{{\tilde X}_i}))},
\end{equation}
where ${P({y_i}|{{\tilde X}_i})}$ denotes the probability that the recovered face ${\tilde X}_i$ is assigned to its identity label $y_i$ by the supervised CNN. So we preserve high-level facial identity information and meanwhile recover low-level structural information of faces. However, we observe that the model produces severe artifacts in recovered face images for fitting to the classification network. Similar to generative adversarial nets (GAN) \cite{R24}, we introduce an adversarial discriminator to alleviated this effect of artifacts. In particular, let $G$ denote the generator modeled by RLA, and $D$ denote the adversarial discriminator modeled by CNN. The optimization procedure can be viewed as a minimax game between $G$ and $D$, where $D$ is trained to discriminate original occlusion-free faces and recovered faces from $G$ through maximizing the log probability of predicting the correct labels (original or recovered) for both of them:
\begin{multline}\label{10}
\mathop {\min }\limits_G \mathop {\max }\limits_D {L_{adv}}(D,G) = \\
\frac{1}{K}\sum\limits_{i = 1}^K {\log (D({X_i}))}  + \log (1 - D(G(X_i^{occ}))),
\end{multline}
while $G$ is trained to recover more real faces which cannot be discriminated by $D$ through minimizing ${L_{adv}}(D,G)$. Both $G$ and $D$ are optimized alternately using stochastic gradient descent as described in \cite{R24}.

We first train the RLA model according to Eqn.~(\ref{8}) by using the multi-stage optimization strategy mentioned previously, and then train the supervised CNN on original occlusion-free face data and the adversarial CNN both on original and recovered faces to obtain a good supervisor and discriminator. In the stage of fine-tuning, we initialize the networks in Fig.~\ref{fig:supervised_learning} using these pre-trained parameters and update the parameters of RLA to optimize the following joint loss function in an end-to-end way:
\begin{equation}\label{11}
L = {L_{mse}} + {L_{sup}} + \mathop {\max }\limits_D {L_{adv}}(D,G).
\end{equation}\label{11}
Here the parameters of the supervised CNN are fixed because it has learned correct filters from original occlusion-free faces. In the other side, we update the parameters of the adversarial CNN to maximize $\mathop {\min }\limits_G {L_{adv}}(D,G)$ in Eqn.~(\ref{10}).

\section{Experiments}

To demonstrate the effectiveness of the proposed model, we evaluate it on two occluded face datasets, in which one contains synthesized occlusion and the other one contains real occlusion. We present qualitative results of occlusion removal as well as quantitative evaluation on face recognition.

\begin{figure*}[t]
	\includegraphics[height=13.3cm]{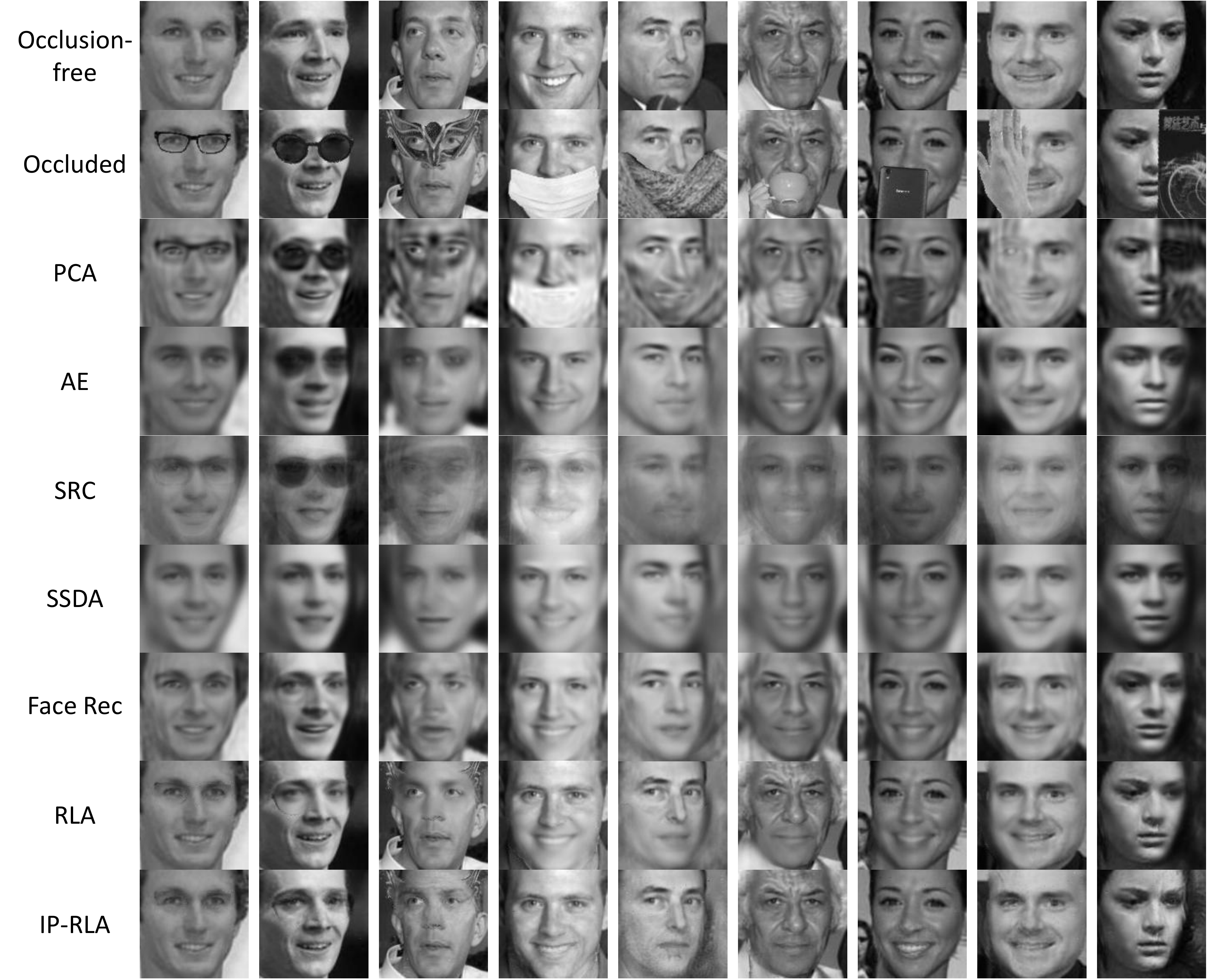}
	\caption{Qualitative comparison of faces recovered by various de-occlusion methods on the occluded LFW dataset. The first row shows occlusion-free faces, the second row shows occluded faces, and the rest rows show the results of PCA, AE, SRC, SSDA, face reconstruction channel of RLA (Face Rec), RLA and IP-RLA, respectively. Best viewed in three times the size.}
	\label{fig:results_lfw1}
\end{figure*}

\subsection{Datasets}

\subsubsection{Training Data} Since it is hard to collect sufficient occluded faces and the corresponding occlusion-free ones in real life to model occluded faces in the wild, we train our model on a synthesized dataset from the CASIA-WebFace dataset \cite{R20}. CASIA-WebFace contains 10,575 subjects and 494,414 face images crawled from the Web. We select around 380,000 near-frontal faces ($-45^{\circ}\sim+45^{\circ}$) from the dataset and synthesize occlusion caused by 9 types of common objects on these faces. The occluding objects we use include glasses, sunglasses, masks, hands, eye masks, scarfs, phones, books and cups. Each type of occluding object has 100 different templates, out of which half are used for generating occlusion on training data and the rest are used for testing data. For each face, we randomly select one template from $9$ types of occlusion to generate the occluded face. Some occlusion templates require a correct location, such as sunglasses, glasses and masks. We add these templates onto specific locations of the faces with reference to detected facial landmarks. The other templates are added onto random locations of the faces to enhance diversity of the produced data. All face images are cropped and coarsely aligned by three key points located at the centers of eyes and mouth, and then resized to $128 \times 128$ gray level ones. Fig.~\ref{fig:occulded_examples} illustrates some examples of occluded faces generated using this approach. We will release the dataset for training upon acceptance.

\subsubsection{Test Data} We use two datasets for testing, i.e., LFW \cite{LFWTech} and 50OccPeople. The latter one is constructed by ourselves. The LFW dataset contains a total of 13,233 face images of 5,749 subjects, which were collected from the Web. Note that LFW does not have any overlap with CASIA-WebFace \cite{R20}. In order to analyze the effects of various occlusion for face recognition, we add all the 9 types of occlusion to every face in the dataset in a similar way for generating training data. Our 50OccPeople dataset contains face images with real occlusion, which contains 50 subjects and 1,200 images. Each subject has one normal face image and 23 face images taken under realistic illumination conditions with the same 9 types of occlusions. The test images are preprocessed by the same way with the training images. It can be seen that both the test datasets have completely different occlusion templates and subjects from the training dataset.

\begin{figure}[t]
	\centering
	\includegraphics[height=5.7cm]{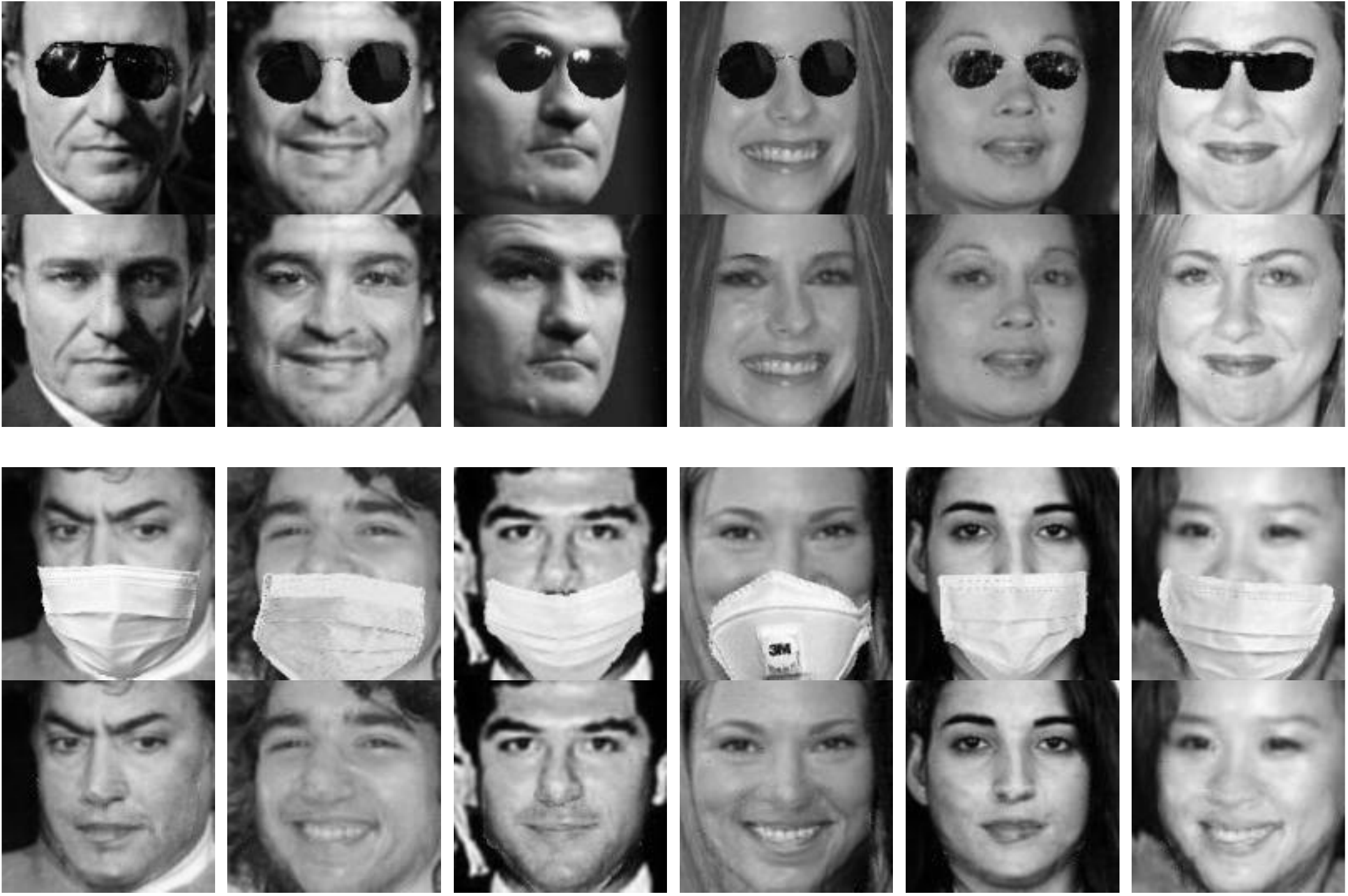}
	\caption{Qualitative comparison on recovered faces of different subjects for the proposed IP-RLA under the condition of the same type of occlusion at the same location on the occluded LFW dataset. The first row shows occluded faces and the second row displays recovered results of IP-RLA.}
	\label{fig:results_diff}
\end{figure}

\begin{figure*}[t]
	\setlength{\abovecaptionskip}{0pt}
	\centering
	\includegraphics[height=11cm]{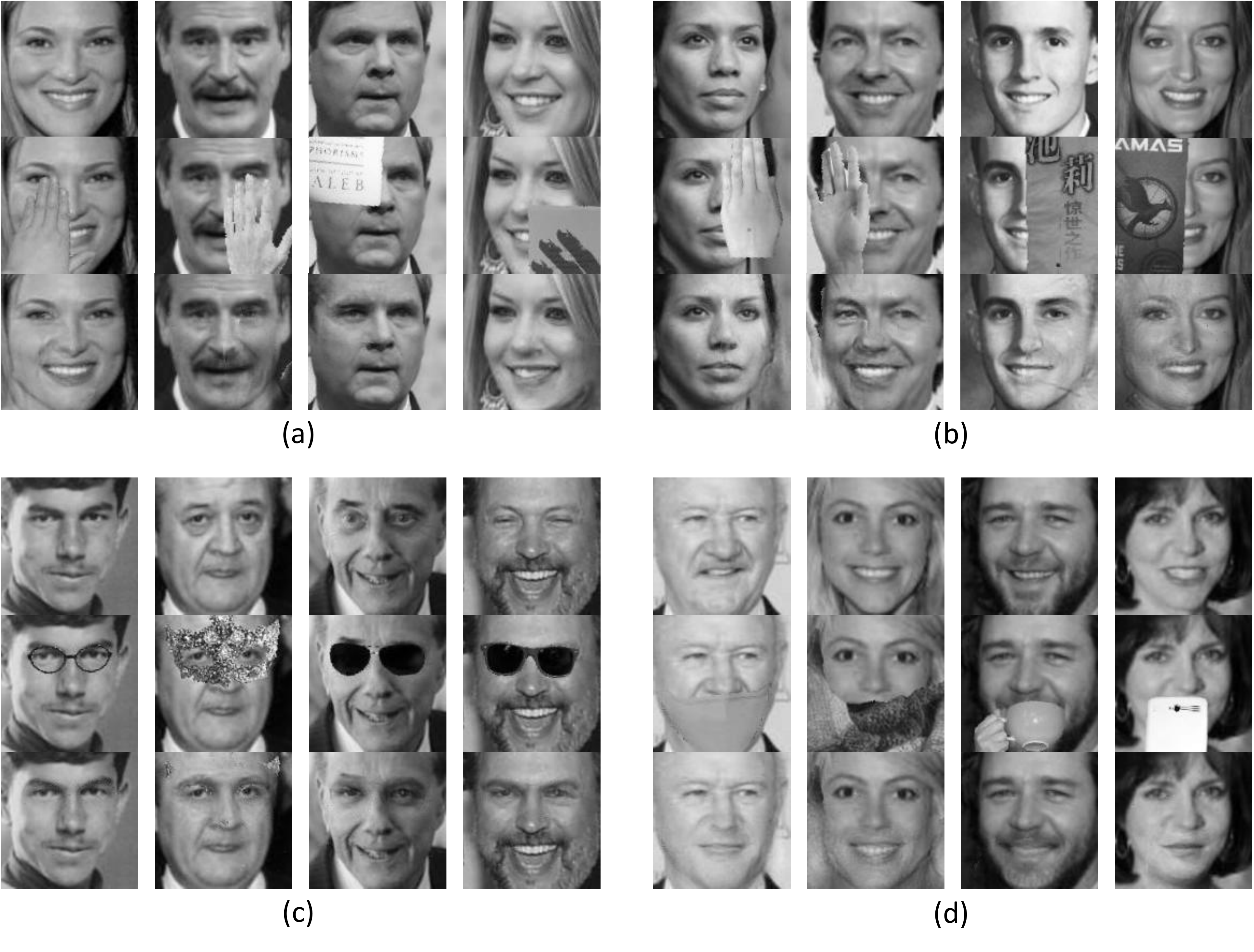}
	\caption{Qualitative results of recovered faces in presence of 4 categories of occlusions for the proposed IP-RLA on the occluded LFW dataset, including (a) quarter of the face at different locations, (b) left or right half of face, (c) upper face and (d) lower face. The first row shows occlusion-free faces, the second row shows occluded faces and the third row displays recovered results of IP-RLA. Note that when eyes or mouths are occluded completely, the restored results may be not very similar to the originals, but it is still possible to correctly predict some general facial attributes, such as genders, sizes, skin colors and expressions.}
	\label{fig:results_lfw2}
\end{figure*}

\subsection{Settings and Implementation Details}

Our model uses a two-layer LSTM network for the encoder and the decoder respectively, and each LSTM has 2,048 hidden units. Each face image is divided into four non-overlapped $64 \times 64$ patches, which is a reasonable size for capturing facial structures and reducing the negative effect of the occlusion. The LSTM encoder reads facial patches from left to right and top to bottom, and meanwhile, the whole image is resized to the same size as a different scale input of the encoder. We set the number of steps of the decoder to 8 for the trade-off between the effectiveness and computational complexity. We use the GoogLeNet \cite{R22} architecture for both the supervised and adversarial CNNs, and the original CASIA-WebFace dataset is used to pre-train the CNNs.

For comparison, a standard autoencoder (AE) with four 2048-dimensional (the same with our model) hidden layers is implemented as a baseline method. We use Principal Component Analysis (PCA) as another baseline, which projects an occluded face image onto a 400 dimensional subspace and then takes PCA reconstruction to be the recovered face. We also include the comparison with Sparse Representation-based Classification (SRC) \cite{R1} and Stacked Sparse Denoising Autoencoder (SSDA) \cite{R15}. We test SRC using a subset of 20K images on CASIA-WebFace. However, even on this sampled training set, the estimation of SRC is already impractically slow. For SSDA, We use the same hyper-parameters with \cite{R15} and the same number and dimensions of hidden layers with our model. In this paper, all the experiments are conducted on a standard desktop with Intel Core i7 CPU and GTX TiTan GPUs.

\begin{figure*}[t]
	\centering
	\includegraphics[height=7.4cm]{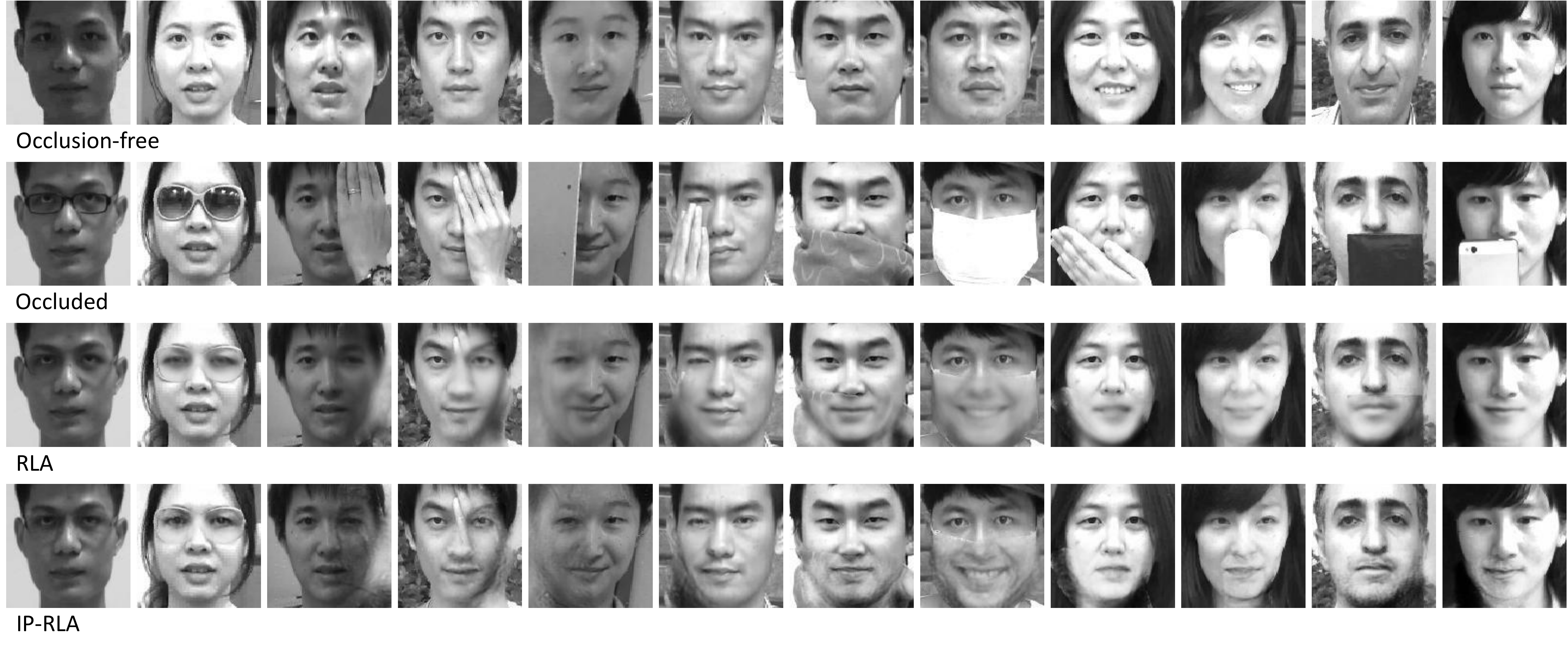}
	\caption{Some examples of recovered faces by RLA and IP-RLA on the 50OccPeople dataset. The first row shows normal faces, the second row shows occluded faces, and the rest rows display the results of RLA and IP-RLA.}
	\label{fig:results_50p}
\end{figure*}

\subsection{Results and Comparisons}

\subsubsection{Occlusion Removal}

We first look into intermediate outputs of the proposed RLA during the process of occlusion removal, which are visualized in Fig.~\ref{fig:results_step}. It can be observed that our model does remove occlusion step by step. Specifically, at the first step, the face reconstruction network of the model produces a probable profile of the face, where occluded parts may not be so clear as non-occluded parts. The occlusion prediction network provides a coarse estimation of the occlusion region. Then the outputs are refined progressively upon the states of previous steps. For example, one can see that more and more structures and textures are added to the face profile, and the shape of the occlusion region becomes sharper and sharper.

To verify the ability of our proposed model for occlusion removal, we present qualitative comparisons with several methods including Principal Component Analysis (PCA), Autoencoder (AE), Sparse Representation-based Classification (SRC) \cite{R1} and Stacked Sparse Denoising Autoencoder (SSDA) \cite{R15}, with different types of occlusion. We also evaluate the contribution of components in our model in ablation study, which include face reconstruction channel of RLA (Face Rec), RLA and the identity-preserving RLA (IP-RLA).

Fig.~\ref{fig:results_lfw1} gives example results of occlusion removal on faces from the occluded LFW dataset. From the figure, one can see that for each type of occlusion, RLA restores occluded parts well and in the meantime retains the original appearance of non-occluded parts. This also demonstrates that the detection of occlusion is rather accurate. Although our model is trained on CASIA-WebFace which has no duplicated subject and occlusion template with the test datasets, our model can still remove occlusion effectively without knowing the type and location of occlusion. Through using the supervised and adversarial CNNs to fine-tune it, the proposed IP-RLA further recovers and sharpens some discriminative patterns, such as edge and texture, in occluded parts. Note that only using the face reconstruction network in the decoder of RLA damages fine-grained structures of non-occluded parts. It is undesired because this might lose key information for the following face recognition task. By comparison, PCA cannot remove occlusion and only make it blur. SRC does not appropriately reconstruct occluded parts and severely damages or changes the appearances of non-occluded parts. AE and SSDA remove occlusion but over-smooths many details, which results in recovered faces biased toward an average face. This clearly demonstrates the advantage of removing occlusion progressively in a recurrent framework.

We also test the proposed IP-RLA on faces of different subjects corrupted by the same type of occlusion at the same location. The results are shown in Fig.~\ref{fig:results_diff}. It can be seen that our method can recover diverse results for different subjects, which demonstrates our method dose not simply produce the mean of occluded facial parts over the training dataset but predicts the meaningful appearances according to non-occluded parts of different subjects.

Furthermore, based on occluded location and area, we divide the occlusion into 4 categories: quarter of the face at different locations, left or right half of the face, upper face and lower face. Fig.~\ref{fig:results_lfw2} compares recovered results of IP-RLA under different occlusion categories on the occluded LFW dataset. As one can see, for quarter of the face, our model can remove it easily. When the left or right half of a face is occluded, although the occluded area is large, our model still produces recovered faces with high similarity to the original occlusion-free faces. The model may exploit facial symmetry and  learn specific feature information from the non-occluded half face. When the upper or lower part of a face is occluded, our model can also remove the occlusion, but the restored parts may be not very similar to the original parts, such as the 4th column in Fig.~\ref{fig:results_lfw2} (c). This is because it is extremely challenging to infer exactly the appearances of the lower (upper) face according to the upper (lower) face. However, it is still possible to correctly predict some general facial attributes, such as genders, sizes, skin colors and expressions.

Besides the synthetic occluded face dataset, we also test our model on the 50OccPeople dataset which is a real occluded face dataset to verify the performance in practice. Some results are illustrated in Fig.~\ref{fig:results_50p}. one can see that our model still obtains good de-occlusion results although it is trained only using synthetic occluded faces.

\begin{table*}[t]
	\begin{center}
		\caption{Equal error rates (EER) of face recognition for occlusion of different types on the occluded LFW dataset.}
		\label{table:results_lfw}
		\begin{tabular*}{0.9\textwidth}{@{\extracolsep{\fill}}cccccccccc}
			\hline\noalign{\smallskip}
			\multicolumn{2}{c}{\multirow{2}{*}{Types of occlusion}} & \multicolumn{3}{c}{Our Model} & \multirow{2}{*}{SSDA} & \multirow{2}{*}{SRC} & \multirow{2}{*}{AE} & \multirow{2}{*}{PCA} & \multirow{2}{*}{\tabincell{c}{Occluded \\ face}} \\
			\cline{3-5}
			\noalign{\smallskip}
			&& IP-RLA & RLA & Face Rec&&&&& \\
			\noalign{\smallskip}
			\hline
			\noalign{\smallskip}
			\multirow{2}{*}{\tabincell{c}{Quarter \\ of face}} & Hand & \textbf{5.6\%} & 6.2\% & 14.8\% & 37.0\% & 40.8\% & 26.8\% & 12.4\% & 6.5\% \\
			& Book & \textbf{5.9\%} & 6.5\% & 15.0\% & 37.7\% & 42.0\% & 27.8\% & 12.5\% & 7.0\% \\
			\hline
			\noalign{\smallskip}
			\multirow{2}{*}{\tabincell{c}{Left/right \\ half of face}} & Hand & \textbf{9.3\%} & 10.5\% & 20.3\% & 39.1\% & 40.1\% & 30.5\% & 18.6\% & 12.8\% \\
			& Book & \textbf{9.8\%} & 11.4\% & 21.4\% & 40.4\% & 43.0\% & 32.6\% & 21.0\% & 13.4\% \\
			\hline
			\noalign{\smallskip}
			\multirow{3}{*}{Upper face} & Glasses & \textbf{5.8\%} & 5.8\% & 12.8\% & 36.3\% & 32.7\% & 25.0\% & 11.8\% & 6.7\% \\
			& Sunglasses & \textbf{9.9\%} & 10.7\% & 22.9\% & 42.9\% & 38.8\% & 33.6\% & 20.7\% & 10.0\% \\
			& Eye mask & \textbf{25.5\%} & 27.3\% & 34.4\% & 44.2\% & 43.2\% & 39.5\% & 33.3\% & 27.2\% \\
			\hline
			\noalign{\smallskip}
			\multirow{3}{*}{Lower face} & Mask & \textbf{9.2\%} & 12.1\% & 20.9\% & 40.3\% & 44.7\% & 31.9\% & 21.4\% & 12.1\% \\
			& Phone & \textbf{7.2\%} & 7.8\% & 15.3\% & 37.8\% & 42.3\% & 28.9\% & 14.2\% & 8.7\% \\
			& Cup & \textbf{5.7\%} & 6.1\% & 15.3\% & 37.9\% & 41.3\% & 28.4\% & 12.7\% & 5.8\% \\
			& Scarf & \textbf{9.3\%} & 11.0\% & 20.9\% & 39.8\% & 44.7\% & 33.7\% & 17.3\% & 10.0\% \\
			\hline
		\end{tabular*}
	\end{center}
\end{table*}

\begin{table*}[t]
	\begin{center}	
		\caption{Average EER of face recognition for all types of occlusion on the 50OccPeople dataset.}
		\label{table:results_50p}
		\begin{tabular*}{0.9\textwidth}{@{\extracolsep{\fill}}cccccccc}
			\hline\noalign{\smallskip}
			\multicolumn{3}{c}{Our Model} & \multirow{2}{*}{SSDA} & \multirow{2}{*}{SRC} & \multirow{2}{*}{AE} & \multirow{2}{*}{PCA} & \multirow{2}{*}{Occluded face} \\
			\cline{1-3}
			\noalign{\smallskip}
			IP-RLA & RLA & Face Rec &&&&& \\
			\noalign{\smallskip}
			\hline
			\noalign{\smallskip}
			\textbf{18.0\%} & 18.2\% & 23.2\% & 42.6\% & 45.5\% & 35.0\% & 25.6\% & 19.1\% \\
			\hline
		\end{tabular*}
	\end{center}
\end{table*}

\subsubsection{Face Recognition}

We carry out the experiment of face verification on the faces recovered by de-occlusion methods to further investigate the ability of our model in recognizing occluded faces. We first extract feature vectors for a pair of face images (one is a occlusion-free face, and the other is a recovered face or an occluded face) and compute the similarity between two feature vectors using Joint Bayesian \cite{R21} to decide whether the pair of faces is from the same subject.

CNN is adopted to extract face features in the experiment. We train a GoogLenet model on CASIA-WebFace, and a 6,144-dimensional feature vector is obtained by concatenating activation outputs of hidden layers before the three loss layers. By reducing dimension using PCA, we have a 800-dimensional feature vector for each face image.

We first evaluate the recognition performance for different types of occlusion on the occluded LFW dataset. We compute the equal error rates (EER) on pre-defined pairs of faces provided by the dataset website. The pair set contains 10 mutually excluded folds, and 300 positive pairs and 300 negative pairs for each fold. Through alternately occluding the two faces in a pair, a total of 12,000 pairs are generated  for testing. Table~\ref{table:results_lfw} reports the verification results for various occlusion and de-occlusion methods. We compare our proposed model with other methods including PCA, AE, SRC and SSDA, and also list the performance of verification on occluded face images for reference. As one can see, the IP-RLA performs better for all types of occlusion as it produces more discriminative occlusion-free faces than other methods. Note that combining with the occlusion detection significantly reduces the error rate compared with recovering faces without using occlusion detection. This is because utilizing occlusion detection to retain non-occluded parts effectively preserve  discriminative information contained in these parts. SRC does not obtain the expected performance as \cite{R1} because the open test set has no identical subject with the training dataset. SSDA performs even worse than the standard Autoencoder (AE), which shows that it cannot handle well the large area of spatially contiguous noise like occlusion although it is effective for removing Gaussian noise and contiguous noise with low magnitude like text. Note that only using face reconstruction (Face Rec) still achieves better performance than the standard Autoencoder (AE). This demonstrates the effectiveness of the progressive recovery framework.

Similar to the observations made in the qualitative analysis, occlusion removal for quarter or left/right half of the face improve better the performance of occluded face recognition because the appearances of occluded facial parts can be predicted according to the non-occluded parts by utilizing facial symmetry. However, recovered faces for upper or lower faces still achieves lower error rate compared with occluded faces, which indicates that our model can learn relations between upper and lower faces and extract discriminative features from non-occluded upper (lower) faces to recover occluded lower (upper) faces.

We also compare the overall verification performance for all types of occlusion on the 50OccPeople dataset. We randomly sample 10,000 pairs (5,000 positive pairs and 5,000 negative pairs) of faces for testing. The EER averaged on all types of occlusion are listed in Table~\ref{table:results_50p}. The verification results shows that our model outperforms other methods and is able to be generalized to real occluded face data.

\section{Conclusions}

In this paper we have proposed a robust LSTM-Autoencoders to address the problem of face de-occlusion in the wild. The proposed model is shown to be able to effectively recover occluded facial parts progressively. The proposed model contains a spatial LSTM network encoding face patches sequentially under different scales for feature representation extraction, and a dual-channel LSTM network decoding the representation to reconstruct the face and detect occlusion step by step. Extra supervised and adversarial CNNs are introduced to fine-tune the robust LSTM autoencoder and enhance the discriminative information about person identity in the recovered faces. Extensive experiments on synthetic and real occlusion datasets demonstrate that the proposed model outperforms other de-occlusion methods in terms of both the quality of recovered faces and the accuracy of occluded face recognition.

\ifCLASSOPTIONcaptionsoff
  \newpage
\fi



\bibliographystyle{IEEEtran}
\bibliography{tip_zf}




\end{document}